\documentclass[11pt]{article}

\usepackage[final]{acl}

\usepackage{times}
\usepackage{latexsym}

\usepackage[T1]{fontenc}

\usepackage[utf8]{inputenc}

\usepackage{microtype}

\usepackage{inconsolata}

\usepackage{graphicx}

\usepackage[utf8]{inputenc} 
\usepackage[T1]{fontenc}    
\usepackage{hyperref}       
\usepackage{url}            
\usepackage{booktabs}       
\usepackage{amsfonts}       
\usepackage{nicefrac}       
\usepackage{microtype}      
\usepackage{xcolor}         
\usepackage{threeparttable}
\usepackage{adjustbox}

\usepackage[dvipsnames]{xcolor}

\usepackage{arydshln}

\usepackage{adjustbox}
\usepackage{wrapfig}

\usepackage{url}
\usepackage[utf8]{inputenc}
\usepackage{booktabs} 
\usepackage{algpseudocode}
\usepackage{xspace}

\definecolor{todo}{rgb}{0.6706,0.5961,0.8667}

\usepackage{xcolor}
\usepackage{url}
\usepackage{enumitem}
\usepackage{pifont}
\definecolor{cite_color}{HTML}{114083}
\definecolor{link_color}{RGB}{153, 0,0} 
\definecolor{url_color}{RGB}{153, 102,  0}
\definecolor{emp_color}{RGB}{0,0,255}

\hypersetup{
  colorlinks=true,
  linkcolor=darkblue,
  citecolor=darkblue,
  urlcolor=darkblue
}

\usepackage{framed}
\definecolor{shadecolor}{rgb}{0.94, 0.97, 1.0}

\usepackage{epsfig}
\usepackage{amsmath}
\usepackage{amssymb}
\usepackage{mathrsfs}
\usepackage{graphicx}
\usepackage{multirow}
\usepackage{booktabs}

\usepackage{wrapfig}
\usepackage{bm}
\usepackage{bbm}
\usepackage{tabularx}

\usepackage{amsfonts}

\usepackage{textcomp}
\usepackage{mathtools}

\usepackage[linesnumbered,ruled,vlined]{algorithm2e}
\SetAlgoNlRelativeSize{0}

\usepackage{amsthm}
\usepackage[capitalize]{cleveref}
 \crefname{section}{Section}{Sections}
 \crefname{theorem}{Theorem}{Theorems}
 \crefname{lemma}{Lemma}{Lemmas}
 \crefname{equation}{Equation}{Equations}
 \crefname{proposition}{Proposition}{Propositions}
 \crefname{claim}{Claim}{Claims}
\crefname{appendix}{Appendix}{Appendices}
   \crefname{algorithm}{Algorithm}{Algorithms}
 \crefname{figure}{Figure}{Figures}
 \crefname{table}{Table}{Tables}
 \crefname{remark}{Remark}{Remarks}
 \crefname{definition}{Definition}{Definitions}
 \crefname{equatinon}{Equation}{Equations}
 \crefname{corollary}{Corollary}{Corollaries}

\allowdisplaybreaks

\usepackage{thmtools}
\usepackage{thm-restate}

\let \oldtextcircled \textcircled
\renewcommand{\textcircled}[1]{\oldtextcircled{\footnotesize #1}}

\setlist[itemize]{leftmargin=9mm}

\crefname{assumption}{Assumption}{Assumptions}

\newcommand{\rave}{\texttt{RAVE}\xspace}
\usepackage{makecell}

\title{RAVE: Re-Allocating Visual Attention in Large Multimodal Models}

\author{
  \textbf{Xi Leng\textsuperscript{1,2}},
  \textbf{Xinhong Ma\textsuperscript{1}},
  \textbf{Ziqiang Dong\textsuperscript{1}},
  \textbf{Feng Zhang\textsuperscript{1,3}},
  \\
  \textbf{Xiaoying Tang\textsuperscript{2}},
  \textbf{Yang Yang\textsuperscript{1}},
  \textbf{Guanjun Jiang\textsuperscript{1}}
  \\
  \textsuperscript{1}Qwen Business Unit of Alibaba,
  \textsuperscript{2}The Chinese University of Hong Kong, Shenzhen,
  \\
  \textsuperscript{3}Beijing Institute of Technology
  \\
  \texttt{\{xinhong.mxh, ziqiang.dzq, chris.yang, guanj.jianggj\}@alibaba-inc.com}
  \\
  \texttt{xileng@link.cuhk.edu.cn, bit\_zhangfeng@bit.edu.cn}
}

\begin{document}

\maketitle

\begin{abstract}

Large multimodal models (LMMs) inherit the self-attention mechanism of pretrained language backbones, yet standard attention can exhibit suboptimal allocation, including cross-modal misallocation between textual and visual evidence and intra-visual imbalance among visual tokens.  We propose \rave{} (Re-Allocating Visual Attention), a lightweight pair-gating mechanism that adds a learned query--key bias to pre-softmax attention scores over visual keys, derived from pre-\textsc{RoPE} query and key features. \rave{} requires no architectural modification to the backbone and can be trained end-to-end with the rest of the model. Across a suite of multimodal benchmarks, \rave{} improves over standard attention by $3$ points on average, with the largest gains on perception-intensive tasks---including multilingual OCR, chart understanding, document VQA, and scene text VQA---where accurate visual grounding is critical.

\end{abstract}

\begin{table*}[t]
\centering
\setlength{\tabcolsep}{4pt}
\renewcommand{\arraystretch}{1.1}
\resizebox{\textwidth}{!}{\begin{tabular}{lccccccc}
\toprule
Method
& \makecell{Training-free}
& \makecell{Extra supervision}
& \makecell{Arch. change}
& \makecell{Input-adaptive}
& \makecell{Cross-modal}
& \makecell{Intra-visual}
& \makecell{Lightweight integration} \\
\midrule
\textsc{AttnReal}~\citep{tu2025attnreal}
&\ding{51} & \ding{55} & \ding{55} & \ding{55} &  \ding{51} & \ding{55} & \ding{51} \\
\textsc{VAR}~\citep{kang2025visualsink}
& \ding{51} & \ding{55} & \ding{55} & \textit{Partial} & \ding{55} & \ding{51} & \ding{51} \\
\textsc{CausalMM}~\citep{zhou2025causalmm}
& \ding{51} & \ding{55} & \ding{55} & \textit{Partial} & \ding{51} & \ding{55} & \ding{51} \\
\textsc{FarSight}~\citep{tang2025farsight}
& \ding{51} & \ding{55} & \ding{55} & \textit{Partial} & \ding{51} & \ding{55} & \ding{51} \\
\midrule
\textsc{D-Attn}~\citep{kuo2025dattn}
& \ding{55} & \ding{55} & \ding{51} & \ding{51} & \ding{51} & \textit{Partial}& \ding{55} \\
\textsc{LLaViT}~\citep{kim2025rethinkingvisualinformationprocessing}
& \ding{55} & \ding{55} & \ding{51} & \ding{51} &  \textit{Partial} & \ding{51} & \ding{55} \\
\midrule
\textsc{TVC}~\citep{sun2025tvc}
& \ding{55} & \ding{51} & \ding{55} & \ding{51}& \ding{51} & \ding{55} & \textit{Partial} \\
\textsc{Look-Back}~\citep{yang2025lookback}
& \ding{55} & \ding{51} & \ding{55} & \ding{51} & \ding{51} & \ding{55} & \textit{Partial} \\
\midrule
\textsc{Gated-Attn}~\citep{qiu2025gatedattention}
& \ding{55} & \ding{55} & \textit{Partial} & \ding{51} & \textit{Partial} & \textit{Partial} & \textit{Partial} \\
\textbf{\rave{} (ours)}
& \ding{55} & \ding{55} & \ding{55} & \ding{51} & \ding{51}& \ding{51} & \ding{51} \\
\bottomrule
\end{tabular}}
\vspace{-5pt}
\caption{Positioning of \rave{} against representative methods for improving visual attention allocation in LMMs.}
\vspace{-10pt}
\label{tab:dual:positioning}

\end{table*}

\section{Introduction}
\label{sec:dual:intro}
A prevalent architectural paradigm for large multimodal models (LMMs), exemplified by the \textsc{LLaVA} family~\citep{liu2023visual,liu2024improved} and the Qwen-VL series~\citep{wang2024qwen2vl,bai2025qwen3vltechnicalreport,bai2025qwen25vltechnicalreport}, projects visual features into the language token space and processes them alongside textual tokens in a unified sequence. This approach relies on the self-attention mechanism inherited from a pretrained language backbone, which we refer to as \textit{standard attention} throughout this work. While simple and effective, this inherited mechanism is not explicitly designed to regulate modality-wise attention allocation. Recent observations suggest that such uniform treatment of visual and textual tokens can lead to suboptimal use of visual evidence during generation, weakening visual grounding in complex multimodal tasks.


This suboptimal allocation manifests at two levels. At the \emph{cross-modal level}, the visual block as a whole may receive insufficient attention relative to textual tokens, reflecting architectural biases inherited from language backbones. This imbalance appears both across decoder depth, where visual tokens become increasingly under-attended in deeper layers~\citep{chen2024fastv,zhang2025moda}, and during decoding, where attention to the visual block decreases as generated answer tokens accumulate~\citep{sun2025tvc,tu2025attnreal,yang2025lookback,tang2025farsight}. At the \emph{intra-visual level}, attention within the visual block may be unevenly distributed: visually uninformative patches can absorb disproportionate mass, forming ``visual sinks''~\citep{kang2025visualsink}, while positional bias may further concentrate attention on specific spatial regions~\citep{kuo2025dattn}. Despite their different manifestations, these phenomena point to a common limitation: standard attention lacks an explicit mechanism for regulating visual attention allocation during multimodal generation. As a result, visual evidence may be under-utilized or misallocated, ultimately undermining semantically precise visual grounding.

Existing approaches mitigate these problems through inference-time attention intervention~\citep{tu2025attnreal,kang2025visualsink,tang2025farsight,zhou2025causalmm}, architectural redesign~\citep{zhang2025layer,kuo2025dattn,kim2025rethinkingvisualinformationprocessing}, or learned visual revisiting~\citep{sun2025tvc,yang2025lookback}. While effective in specific scenarios, they often rely on hand-crafted intervention heuristics with limited context awareness, incur structural departures from standard Transformer attention that may complicate integration, or require auxiliary training pipelines involving curated data and specialized generation protocols. There remains a need for a minimalist, drop-in mechanism that enables context-aware recalibration of attention allocation, while remaining compatible with standard training and inference workflows.

To this end, we propose \rave{} (\textbf{R}e-\textbf{A}llocating \textbf{V}isual \textbf{A}ttention), a lightweight pair-gating mechanism that corrects standard attention in a drop-in manner. \rave{} adds a learned query--key bias to the pre-softmax attention scores over visual keys, derived from pre-\textsc{RoPE} query and key features. This enables context-adaptive recalibration of visual attention within the native attention operator, without altering the backbone architecture or disrupting standard end-to-end training.

We evaluate \rave{} on a diverse suite of multimodal benchmarks covering both perception-intensive and reasoning-oriented tasks. \rave{} consistently improves over standard attention, achieving an average gain of $3$ points. The improvements are particularly pronounced on perception-intensive tasks, including multilingual OCR, chart understanding, document understanding, and text-rich VQA, where accurate visual grounding is critical. These results show the effectiveness of context-aware attention recalibration within standard attention for improving multimodal performance.

Our contributions are threefold:
\begin{itemize}[leftmargin=1.2em,labelsep=0.4em,itemsep=1pt,topsep=2pt,parsep=0pt,partopsep=0pt]

    \item We revisit standard attention in LMMs and identify suboptimal visual attention allocation as a key bottleneck for effective visual grounding in multimodal generation.

    \item We propose \rave{}, a lightweight, architecture-preserving pair-gating mechanism that modulates pre-softmax attention scores over visual keys with negligible additional parameters.

    \item We demonstrate that \rave{} consistently outperforms standard attention across a wide range of multimodal benchmarks, achieving 3-point average gains, with particularly strong improvements on perception-intensive tasks.

\end{itemize}

\section{Related Work}
\label{sec:related}

\textbf{Attention allocation in large multimodal models.} Modern large multimodal models (LMMs), including the \textsc{LLaVA} family and the Qwen-VL series, typically project visual features into the language token space and process them together with textual tokens in a unified sequence using a decoder-style transformer. Under this paradigm, the inherited self-attention mechanism serves as the default interface for multimodal fusion, making visual grounding highly dependent on how attention is allocated across visual and textual tokens during generation. As a result, attention allocation has become an important lens for understanding and improving multimodal reasoning in LMMs.

\noindent
\textbf{Visual-attention misallocation in LMMs.}
Recent studies have identified two complementary forms of suboptimal visual attention allocation. \emph{\textbf{Cross-modal imbalance}: visual tokens, as a block, are systematically under-attended relative to textual tokens.}
This imbalance manifests along two axes. \emph{Across decoder depth,} \citet{chen2024fastv} show on \textsc{LLaVA}-class models that, despite occupying a substantial fraction of the input sequence, visual tokens receive significantly less attention than textual tokens, with the disparity increasing in deeper layers. \citet{zhang2025moda} attribute this effect to a combination of language-side bias in attention scores and layer-wise decay in activation magnitude, and show that such imbalance can severely degrade fine-grained perception. \emph{During decoding,} a temporally indexed counterpart has been reported under different names: \citet{sun2025tvc} describe ``visual forgetting,'' where attention to the image decays as long-chain reasoning progresses; \citet{tu2025attnreal} observe that historical output tokens increasingly dominate attention as generation unfolds; and \citet{yang2025lookback} report that visual attention can diminish to near zero in later reasoning stages. \citet{tang2025farsight} further relate this temporal decay to the distance-dependent attenuation induced by rotary position embeddings~\citep{su2024roformer}. Collectively, these studies suggest that visual attention can be systematically disadvantaged by architectural and positional biases in standard attention, which can compromise reliable visual grounding and weaken fine-grained perception.
\emph{\textbf{Intra-visual imbalance}: within the visual block, attention is unevenly distributed across visual tokens.}
This imbalance also manifests in two forms. First, \citet{xiao2023streamingllm} identify ``attention sinks'' in language models, where a small number of tokens absorb disproportionate attention regardless of semantic relevance. \citet{kang2025visualsink} show that an analogous phenomenon arises in LMMs, where attention concentrates on visually uninformative tokens such as corner patches or homogeneous background regions. \citet{tang2025farsight} further describe a related ``attention collapse,'' in which outlier tokens—arising from either modality—receive disproportionately high attention scores. Second, \citet{kuo2025dattn} identify a position-induced bias within the visual block: with row-major flattened image tokens, positional encoding can cause textual queries to favor spatially proximate patches in the 1D sequence, leading to attention that is spatially biased rather than semantically grounded. Consequently, even when nontrivial attention is assigned to the visual block, a substantial portion of that mass may be allocated to uninformative or systematically biased regions.


Together, these cross-modal and intra-visual imbalances point to a key limitation of standard attention: it lacks an explicit mechanism for regulating visual attention allocation during multimodal generation, a gap that \rave{} aims to bridge.

\begin{figure*}[t]
  \centering
  \includegraphics[width=0.49\textwidth]{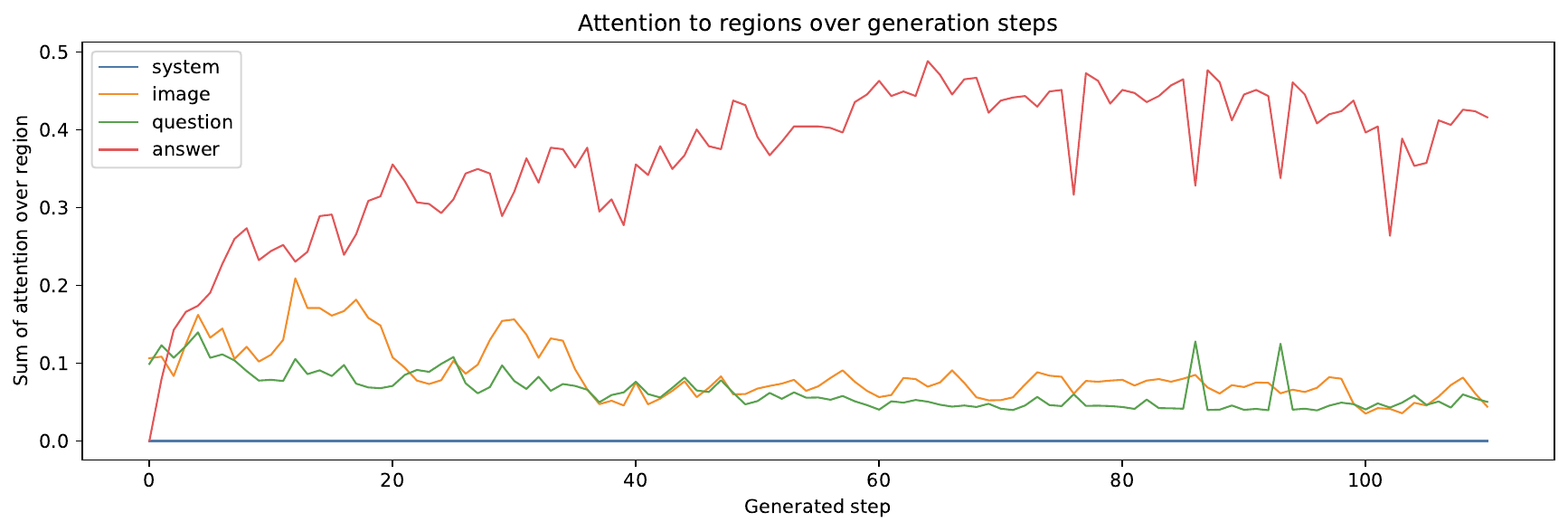}\hfill
  \includegraphics[width=0.49\textwidth]{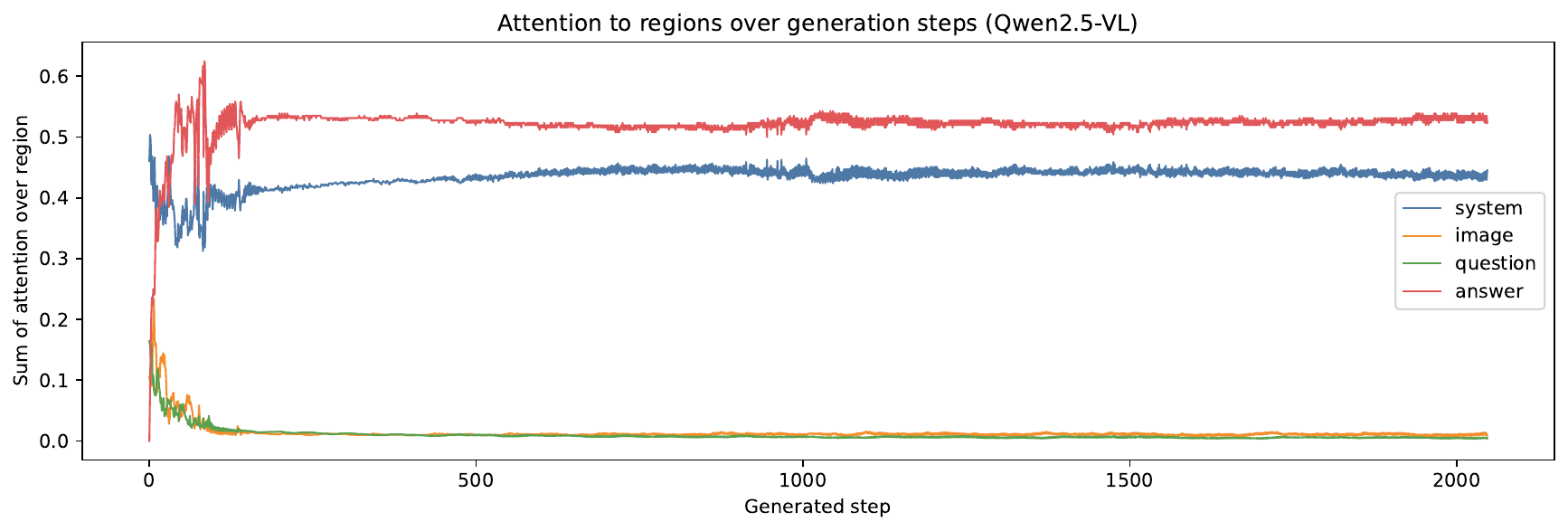}
  \vspace{-5pt}
  \caption[Attention mass over four input segments in two Qwen-VL models.]{Attention mass assigned by an answer token to the four input segments---\emph{system}, \emph{image}, \emph{question}, and \emph{answer}---as generation proceeds, averaged over decoder layers and attention heads. \textbf{Left}: Qwen3-VL-4B-Instruct, which does not apply a default system prompt, so the \emph{system} mass is identically zero. \textbf{Right}: Qwen2.5-VL-7B-Instruct with its default system prompt \texttt{``You are a helpful assistant''}. }
  \label{fig:dual:trend-cases}
  \vspace{-10pt}
\end{figure*}

\vspace{2pt}
\noindent
\textbf{Mitigating visual-attention misallocation in LMMs.}
Existing mitigation strategies can be grouped into three families, each operating at a different place in the pipeline. 
Table~\ref{tab:dual:positioning} summarizes representative methods along the axes most relevant to our discussion.
\emph{\textbf{Training-free methods intervene at inference time.}} \textsc{AttnReal}~\citep{tu2025attnreal} recycles excessive attention from historical output tokens back to visual tokens; 
\textsc{VAR}~\citep{kang2025visualsink} identifies visual sink tokens and redistributes their surplus attention toward more informative visual regions, reducing intra-visual attention bias; 
\textsc{CausalMM}~\citep{zhou2025causalmm} applies causal adjustment and counterfactual attention interventions to mitigate modality-prior-induced hallucinations; \textsc{FarSight}~\citep{tang2025farsight} modifies the causal mask to absorb attention from outlier tokens and uses a progressively diminishing mask schedule to counter RoPE-induced distance-dependent attenuation during decoding. Although these methods require no additional training, they typically rely on hand-crafted intervention rules rather than learned, input-adaptive corrections. \emph{\textbf{Architectural methods redesign multimodal attention.}} \textsc{D-Attn}~\citep{kuo2025dattn} decomposes self-attention into modality-specific paths (vision-to-vision, text-to-vision, text-to-text) and modifies the positional treatment of visual tokens; \textsc{LLaViT}~\citep{kim2025rethinkingvisualinformationprocessing} learns separate $\{Q,K,V\}$ projections for visual tokens and enables bidirectional attention within the visual block; \textsc{AKI}~\citep{wang2025aki} relaxes the causal mask to allow visual tokens to attend to subsequent question tokens. These methods can be effective, but they often introduce additional architectural complexity or require non-trivial re-alignment with existing LMM checkpoints.
\emph{\textbf{Training-based revisiting methods learn explicit visual reactivation behaviors.}} \textsc{TVC}~\citep{sun2025tvc} trains the model to re-attend to visual input by inserting visual embeddings and bridging prompts into long-chain reasoning data, and further applies inference-time calibration by periodically reintroducing compressed visual tokens; \textsc{Look-Back}~\citep{yang2025lookback} uses supervised fine-tuning and reinforcement learning to elicit a special $\langle\text{back}\rangle$ token that redirects attention to the image. These methods can improve visual utilization, but they require auxiliary training pipelines with curated reasoning data and specialized generation protocols.

\rave{} fills a distinct design space: a lightweight, learnable module placed directly inside the native attention. While it shares the gating philosophy of recent works such as \textsc{Gated-Attn}~\citep{qiu2025gatedattention}, \rave{} shifts the intervention from the attention output to the pre-softmax attention scores associated with visual keys. This shift is fundamental: whereas \textsc{Gated-Attn} modulates how attended features are integrated, \rave{} modulates where attention is allocated. By directly recalibrating the competition for attention mass, \rave{} provides a mechanism for regulating visual attention allocation that is relevant to both cross-modal and intra-visual misallocation, while remaining natively compatible with standard LMM training and inference.


\section{Preliminaries}
\label{sec:dual:prelim}
To analyze attention allocation in decoder-only LMMs, we adopt segment-wise attention mass, a commonly used diagnostic statistic for characterizing token-type attention patterns and visual-token utilization~\citep{chen2024fastv,tu2025attnreal,yang2025lookback}. Following this line of analysis, we instantiate the statistic under the segment partition used in this work. Specifically, we separate the context into system-prompt tokens $\mathcal{I}_\text{sys}$ introduced by the chat template, visual tokens $\mathcal{I}_\text{img}$ produced by the vision encoder, question tokens $\mathcal{I}_\text{que}$ from the user instruction, and answer tokens $\mathcal{I}_\text{ans}$ generated autoregressively by the model. We write $N$ for the total sequence length after generation and let $\mathcal{I}_s \subseteq \{1,\dots,N\}$ denote the set of token positions belonging to segment $s \in \{\text{sys},\text{img},\text{que},\text{ans}\}$. At layer $\ell$ and head $h$, standard attention with rotary position embedding~\citep{su2024roformer} computes:

\begingroup
\small
\setlength{\jot}{1pt}
\begin{align}
  Q^{\ell,h} 
  &= \mathrm{RoPE}(\bar{Q}^{\ell,h}),\;
  K^{\ell,h} 
  = \mathrm{RoPE}(\bar{K}^{\ell,h}),
  \label{eq:dual:rope} \\
  L^{\ell,h}_{ij} 
  &= \frac{\langle Q^{\ell,h}_i,K^{\ell,h}_j\rangle}{\sqrt{d_k}},\;
  A^{\ell,h}_{ij} 
  = \operatorname*{softmax}_{j}(L^{\ell,h}_{ij}\!+\!M_{ij})
  \label{eq:dual:std-attn}
\end{align}
\endgroup

\noindent
where $\bar Q,\bar K\in\mathbb{R}^{N\times d_k}$ denote pre-\textsc{RoPE} query and key projections; $M$ is the causal mask; $L$ and $A$ are the attention scores and probabilities. With grouped-query attention~\citep{ainslie2023gqa}, multiple query heads share one key/value head; thus, throughout this work, ``head'' refers to a query head, and the gating module is instantiated at the query-head granularity in Section~\ref{sec:dual:method}.

We characterize attention allocation across input segments from two complementary views. For an answer position $t \in \mathcal{I}_\text{ans}$, we first compute the \emph{segment-wise layer-averaged attention mass}:
\begin{equation}
\begin{aligned}
  \alpha^{(t)}_{s} \;&=\; \frac{1}{HL}\sum_{\ell=1}^{L}\sum_{h=1}^{H}\sum_{j\in\mathcal{I}_s} A^{\ell,h}_{tj},\\ &s\in\{\text{sys},\text{img},\text{que},\text{ans}\},
  \end{aligned}
  \label{eq:dual:vis-mass}
\end{equation}
which aggregates, over all decoder layers $\ell$ and heads $h$, the attention mass assigned by token $t$ to segment $s$. By construction, $\sum_{s}\alpha^{(t)}_{s}=1$ at each decoding step, so $\{\alpha^{(t)}_{s}\}_{s}$ can be interpreted as a distribution of attention mass over input segments.

To examine how attention allocation varies across decoder layers, we also compute a \emph{layer-resolved} variant:
\begin{equation}
  \alpha^{(t),\ell}_{s} \;=\; \frac{1}{H}\sum_{h=1}^{H}\sum_{j\in\mathcal{I}_s} A^{\ell,h}_{tj},
  \label{eq:dual:vis-mass-layer}
\end{equation}
which preserves the layer dimension while using the same segment partition. 

\section{Motivation}
\label{sec:dual:motivation}

 \paragraph{Setup.}

We analyze the attention allocation dynamics of Qwen3-VL-4B-Instruct~\citep{bai2025qwen3vltechnicalreport} and Qwen2.5-VL-7B-Instruct~\citep{bai2025qwen25vltechnicalreport}. To cover different visual grounding scenarios, we use representative probe inputs spanning \emph{visual perception} tasks (e.g., dense OCR and chart understanding) and \emph{visual--textual reasoning} tasks (e.g., mathematical and physics diagrams). For each example, we track both the segment-wise layer-averaged mass $\alpha^{(t)}_s$ and the layer-resolved mass $\alpha^{(t),\ell}_s$ throughout decoding.

\begin{figure*}[t]
  \centering
  \includegraphics[width=0.49\textwidth]{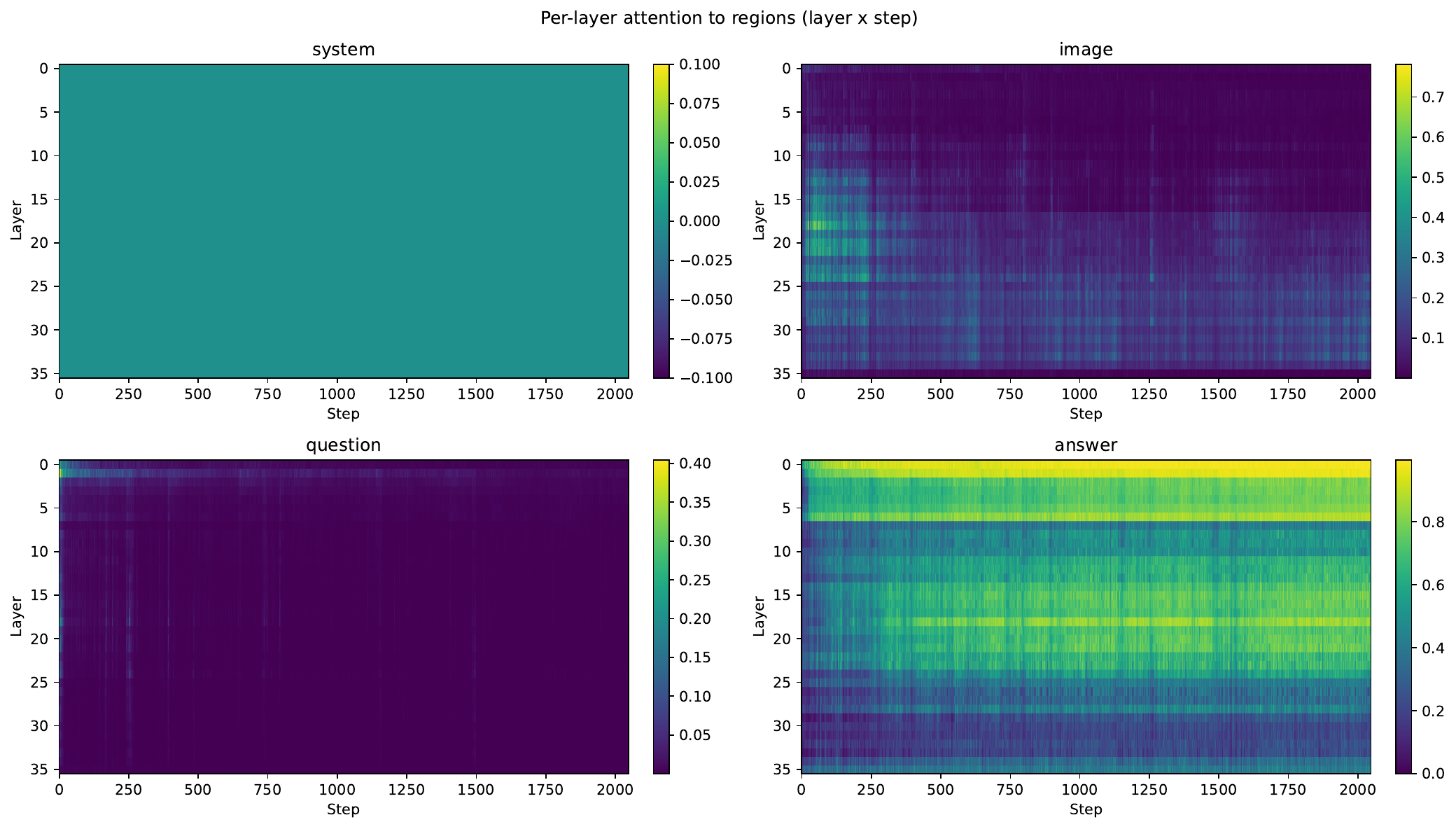}\hfill
  \includegraphics[width=0.49\textwidth]{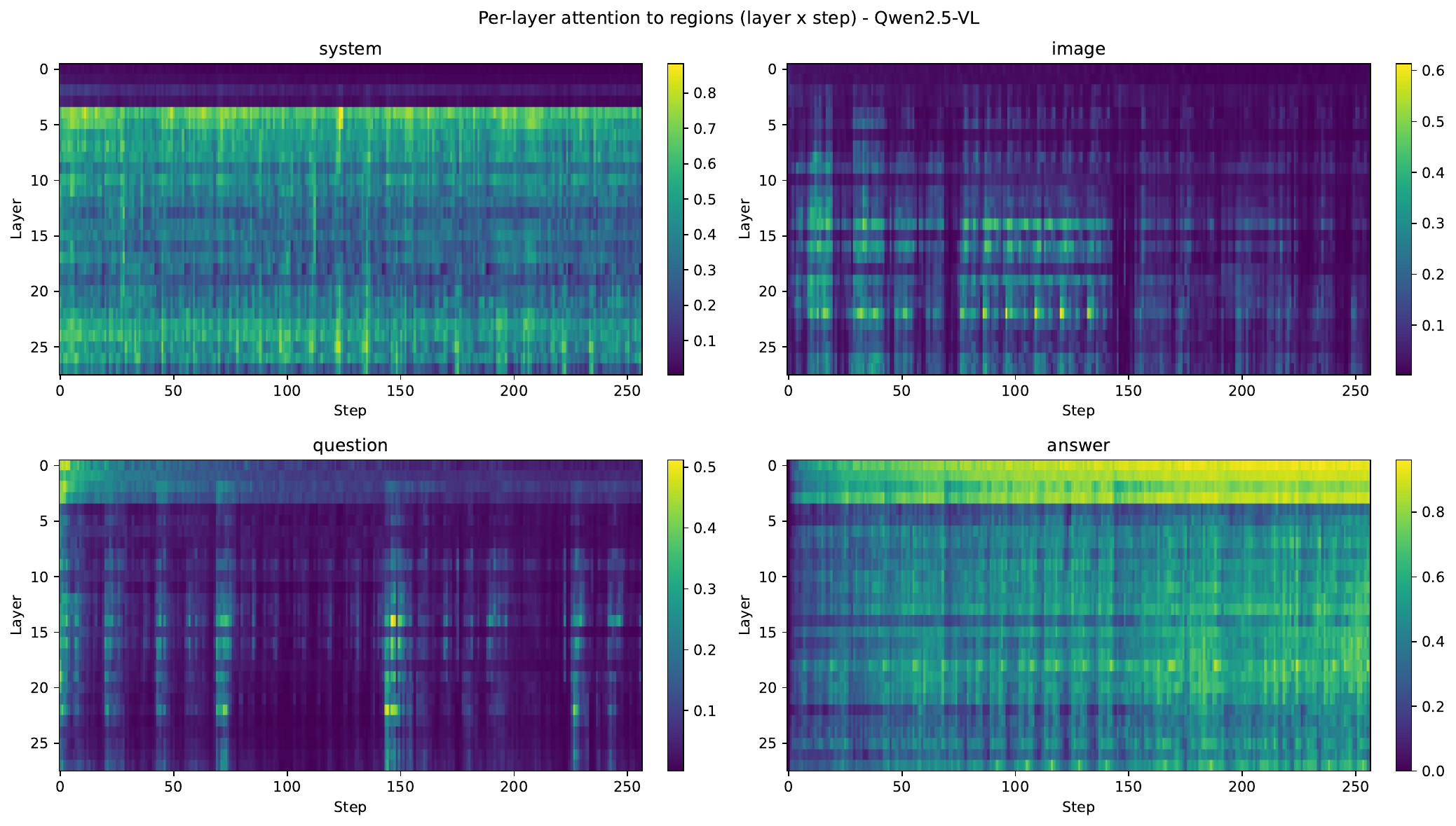}
  \vspace{-5pt}
  \caption[Per-layer attention mass over input segments in two Qwen-VL models.]{Per-layer attention mass (layer $\times$ decoding step) for Qwen3-VL-4B-Instruct (left) and Qwen2.5-VL-7B-Instruct (right). Each panel shows, across layers and decoding steps, the total attention mass assigned by the currently generated token to the four input segments: \emph{system}, \emph{image}, \emph{question}, and \emph{answer}.}
  \label{fig:dual:heatmap}
  \vspace{-10pt}
\end{figure*}

\noindent
\paragraph{Observation 1: Standard attention exhibits progressive visual-attention dilution.} Figure~\ref{fig:dual:trend-cases} shows the segment-wise layer-averaged attention mass $\alpha^{(t)}_s$ on a symbolic mathematical reasoning example for Qwen3-VL-4B-Instruct and Qwen2.5-VL-7B-Instruct. In both models, attention allocation shifts as decoding proceeds: visual-token mass starts at a non-negligible level but steadily declines, whereas answer-token mass grows in parallel. This suggests a gradual transition from visually grounded generation to increasingly self-conditioned generation based on the model’s own textual history. The trend is consistent with prior reports of visual-attention dilution in different settings, including mathematical reasoning~\citep{sun2025tvc}, hallucination snowballing~\citep{tang2025farsight}, and fine-grained cognition and emotion understanding~\citep{zhang2025moda}, and further indicates that visual under-allocation remains a recurring issue in recent decoder-only LMMs.

The right panel of Figure~\ref{fig:dual:trend-cases} illustrates a negative case associated with visual-attention dilution. In Qwen2.5-VL-7B-Instruct, visual attention decreases during generation, while nontrivial attention remains on the default system prompt \texttt{``You are a helpful assistant''}, which carries little task-specific content. This prompt-side concentration resembles attention-sink behavior in language models~\citep{xiao2023streamingllm}. The model consequently produces unstable image-conditioned reasoning and repeats until reaching the decoding limit. By contrast, Qwen3-VL-4B-Instruct, which does not use a default system prompt in our setting, avoids this prompt-side concentration but still shows declining visual attention.


\noindent
\paragraph{Observation 2: The dilution pattern is visible across decoder layers.}
The same pattern is also visible in the layer-resolved view. Figure~\ref{fig:dual:heatmap} presents $\alpha^{(t),\ell}_s$ on a representative OCR-style input for the same two backbones. Despite being substantially different from the mathematical reasoning example in Figure~\ref{fig:dual:trend-cases}, it exhibits a similar trend during decoding: visual-token mass decreases, while answer-token mass increases. The layer-resolved view further shows that this shift is a decoding-stage phenomenon that persists throughout the decoder rather than a depth-specific effect. These observations suggest that visual-attention dilution is not merely a task-specific artifact of one example, but a recurring allocation pattern in standard attention.

\section{Method}
\label{sec:dual:method}


Motivated by the visual-attention dilution observed in Section~\ref{sec:dual:motivation} and the intra-visual attention imbalance discussed in Section~\ref{sec:related}, we propose \rave{} (\textbf{R}e-\textbf{A}llocating \textbf{V}isual \textbf{A}ttention), a lightweight pair-gating mechanism that serves as a drop-in correction to standard attention in decoder-only LMMs.

\subsection{Pair gate from pre-\textsc{RoPE} query and key}
\label{sec:dual:module}

Let $\bar Q^{\ell,h}, \bar K^{\ell,h}\in\mathbb{R}^{N\times d_k}$ denote the \emph{pre-rotary} query and key features of layer $\ell$ and head $h$, i.e., the inputs to $\mathrm{RoPE}(\cdot)$ in Eq.~\eqref{eq:dual:rope}. We derive the gating signal from these pre-\textsc{RoPE} representations rather than from the post-\textsc{RoPE} $Q$ and $K$. This choice is motivated by recent analyses showing that rotary position encoding can introduce systematic bias into cross-modal attention~\citep{zhao2025attention,tang2025farsight,kuo2025dattn}. Using $\bar Q$ and $\bar K$ therefore helps decouple the gating signal from positional decay, allowing it to depend primarily on modality-conditioned content features.

At each layer $\ell$, we attach a pair of learnable linear projections
\begin{equation}
  W^{\ell}_{gq},\, W^{\ell}_{gk}\in\mathbb{R}^{d_k\times 1},
\end{equation}
shared across all $H$ query heads in that layer. For each head $h$, these projections produce per-token scalar scores
\begin{equation}
  s^{\ell,h}_{q} = \bar Q^{\ell,h} W^{\ell}_{gq}\in\mathbb{R}^{N},
  \;
  s^{\ell,h}_{k} = \bar K^{\ell,h} W^{\ell}_{gk}\in\mathbb{R}^{N}.
  \label{eq:dual:scores}
\end{equation}
We then form a pairwise gate by taking the outer product of these scores:
\begin{equation}
  G^{\ell,h}_{ij} = \phi\!\left(s^{\ell,h}_{q,i}\, s^{\ell,h}_{k,j}\right),
  \label{eq:dual:pairgate}
\end{equation}
where $\phi$ maps the pairwise score into the gating bias used for visual attention recalibration. Since each entry $G^{\ell,h}_{ij}$ is computed from the current query token and a candidate key token, the gate provides a context-aware correction for each query--key pair. The projections are shared within each layer, introducing only negligible parameter overhead. This design preserves the original attention structure of the backbone and can therefore be seamlessly integrated into a wide range of decoder-only LMMs.

\subsection{Pre-softmax additive recalibration over visual keys}

Based on the standard attention logits $L^{\ell,h}_{ij}$ defined in Eq.~\eqref{eq:dual:std-attn}, RAVE injects the pairwise gate $G^{\ell,h}_{ij}$ into the pre-softmax scores associated with visual keys:
\begin{equation}
\widetilde{L}^{\ell,h}_{ij}
=
L^{\ell,h}_{ij}
+ G^{\ell,h}_{ij}\cdot \mathbbm{1}[j\in\mathcal{I}_\mathrm{img}],
\label{eq:dual:rave-logit}
\end{equation}
where $\mathbbm{1}[\cdot]$ is the indicator function. The recalibrated attention distribution is then computed as
\begin{equation}
\widetilde{A}^{\ell,h}_{ij}
=
\mathrm{softmax}_{j}
\left(
\widetilde{L}^{\ell,h}_{ij}
+
M_{ij}
\right),
\label{eq:dual:rave-attn}
\end{equation}
where $M_{ij}$ denotes the causal attention mask. Since the correction is added before normalization, RAVE recalibrates how visual and textual keys compete for attention probability mass within the native attention operator, rather than reweighting an already normalized attention distribution. As this correction is computed at the query--key-pair level for visual keys, it also enables redistribution within the visual-token set, providing a direct handle on intra-visual attention imbalance. Restricting the correction to visual-key positions keeps text-key logits intact, thereby limiting direct interference with the pretrained language-side attention prior.

\subsection{Prefill-aware and GQA-aware partial-head application}
\rave{} recalibrates attention to visual keys during autoregressive decoding for each answer-position query, modulating both the strength of visual evidence incorporation and its distribution across visual tokens. The same recalibration is also applied during prefill, where question tokens adjust their attention to visual keys and refine the image-conditioned representations that subsequent answer generation builds upon.

Since \rave{} is introduced into decoder-only LMMs built on a pretrained language backbone, we use a partial-head intervention to avoid over-disrupting the inherited language prior. This partition is implemented in a GQA-aware manner. Modern decoder-only LMMs commonly adopt grouped-query attention (GQA), where $H_q$ query heads share $H_{kv}$ key/value heads. We refer to the query heads that share the same key/value head as one GQA group. Let $r=H_q/H_{kv}$ denote the number of query heads in each group. For each GQA group, we select a fraction $p$ of its $r$ query heads for \rave{} and leave the remaining query heads unchanged.

Let $\mathcal{H}_\mathrm{rave}$ denote the selected query-head subset, with $|\mathcal{H}_\mathrm{rave}|=pH_q$. For heads in $\mathcal{H}_\mathrm{rave}$, we use the recalibrated logits $\widetilde{L}^{\ell,h}_{ij}$; for the remaining heads, we keep the original logits $L^{\ell,h}_{ij}$:
\begin{equation}
\widehat{L}^{\ell,h}_{ij}
=
\begin{cases}
\widetilde{L}^{\ell,h}_{ij}, & h\in\mathcal{H}_\mathrm{rave},\\
L^{\ell,h}_{ij}, & h\notin\mathcal{H}_\mathrm{rave}.
\end{cases}
\label{eq:dual:head-partition}
\end{equation}
We use $p=0.25$ in our main configuration and study the effect of different head ratios in Section~\ref{sec:dual:experiments}.

\begin{algorithm}[t]
\small
\caption{Forward pass of \rave{} attention for a selected query head $h\in\mathcal{H}_\mathrm{rave}$; ungated heads use vanilla attention.}
\label{alg:dual}
\KwIn{Pre-\textsc{RoPE} query/key features $\bar Q^{\ell,h},\bar K^{\ell,h}\in\mathbb{R}^{N\times d_k}$; values $V^{\ell,h}\in\mathbb{R}^{N\times d_v}$; visual-token index set $\mathcal{I}_\mathrm{img}$; layer-shared gate weights $W^{\ell}_{gq}, W^{\ell}_{gk}\in\mathbb{R}^{d_k\times 1}$; causal mask $M$.}
\KwOut{Output features $O^{\ell,h}\in\mathbb{R}^{N\times d_v}$.}

$Q^{\ell,h}\leftarrow\mathrm{RoPE}(\bar Q^{\ell,h}),\quad
K^{\ell,h}\leftarrow\mathrm{RoPE}(\bar K^{\ell,h})$\;

$L^{\ell,h}\leftarrow Q^{\ell,h}(K^{\ell,h})^\top/\sqrt{d_k}$\;

$s_q^{\ell,h}\leftarrow \bar Q^{\ell,h} W^{\ell}_{gq},\quad
s_k^{\ell,h}\leftarrow \bar K^{\ell,h} W^{\ell}_{gk}$\;

$G^{\ell,h}\leftarrow \phi\!\left(s_q^{\ell,h}(s_k^{\ell,h})^\top\right)$\;

$\widetilde L^{\ell,h}\leftarrow L^{\ell,h}$\;

$\widetilde L^{\ell,h}_{:,\mathcal{I}_\mathrm{img}}
\leftarrow
L^{\ell,h}_{:,\mathcal{I}_\mathrm{img}}
+
G^{\ell,h}_{:,\mathcal{I}_\mathrm{img}}$\;

$\widetilde A^{\ell,h}\leftarrow\mathrm{softmax}_j(\widetilde L^{\ell,h}+M)$\;

$O^{\ell,h}\leftarrow \widetilde A^{\ell,h}V^{\ell,h}$\;

\Return $O^{\ell,h}$\;
\end{algorithm}

\begin{table*}[t]
\centering
\setlength{\tabcolsep}{3.2pt}
\renewcommand{\arraystretch}{1.08}
\resizebox{\textwidth}{!}{
\begin{tabular}{lccccccccc|ccc}
\toprule
 \multirow{2}{*}{Attention Method}
& \multicolumn{6}{c}{Perception-intensive}
& \multicolumn{3}{c}{Reasoning / Knowledge-heavy}
& \multirow{2}{*}{Perc. Avg.}
& \multirow{2}{*}{Reason. Avg.}
& \multirow{2}{*}{Overall Avg.} \\
\cmidrule(lr){2-7}
\cmidrule(lr){8-10}

& MTVQA
& ChartQA
& DocVQA
& OCRBench
& InfoVQA
& TextVQA
& MathVista
& MMMU
& MMMU-Pro
& & & \\
\midrule
 \textit{Standard Attention}
& 6.51 & 23.20 & 35.85 & 40.60 & 24.86 & 62.17
& 24.60 & 30.56 & 20.00 & 32.20 & 25.05 & 29.82 \\
\textsc{VAR}
& 6.46 & 22.60 & 35.56 & 40.10 & 24.29 & 62.39
& 24.50 & 31.00 & 19.03 & 31.90 & 24.84 & 29.55 \\
\textsc{D-Attn}
& 6.23 & 23.92 & 35.14 & 39.40 & 26.21 & 62.18
& \textbf{27.90} & \textbf{39.33} & \textbf{20.29}
& 32.18 & \textbf{29.17} & 31.18 \\
\textsc{Gated-Attn}
& 6.05 & 20.92 & 33.61 & 37.70 & 24.22 & 60.54
& 24.80 & 31.16 & 15.83 & 30.51 & 23.93 & 28.31 \\
\rave{}
& \textbf{9.50} & \textbf{26.28} & \textbf{43.55} & \textbf{44.20} & \textbf{28.21} & \textbf{65.90}
& 25.40 & 32.22 & 20.16
& \textbf{36.26} & 25.93 & \textbf{32.81} \\
\bottomrule
\end{tabular}}
\vspace{-5pt}
\caption{
Main results under the controlled Mistral-v0.3-7B setting. The trainable variants use matched model and training configurations, whereas \textsc{VAR} is applied post hoc to the same trained \textit{Standard Attention} checkpoint. Benchmarks are grouped into perception-intensive and reasoning/knowledge-heavy tasks.
}
\label{tab:dual:main-results}
\vspace{-10pt}
\end{table*}

\subsection{The \rave{} algorithm}
\label{sec:dual:algorithm}

We initialize $W^{\ell}_{gq}$ and $W^{\ell}_{gk}$ near zero and use $\tanh$ for $\phi$ in \cref{eq:dual:pairgate}, so that $G^{\ell,h}_{ij}\approx 0$ at the beginning of training and the pretrained backbone behavior is preserved. Since \rave{} adds only two layer-wise scalar projections shared across query heads, its parameter overhead is $2Ld_k$, negligible compared with the backbone size.

Algorithm~\ref{alg:dual} summarizes the forward pass of \rave{} for a selected gated query head. During training, the gate parameters are optimized jointly with the LMM under the standard next-token cross-entropy loss, without auxiliary supervision or an additional gating objective. Overall, \rave{} provides an end-to-end trainable mechanism for visual attention recalibration, enabling context-aware modulation that helps mitigate both cross-modal and intra-visual attention misallocation, while keeping the backbone architecture and standard LMM training pipeline unchanged.

\section{Experiments}
\label{sec:dual:experiments}

\subsection{Setup}

\paragraph{Experimental protocol and baselines.}
Following the controlled protocol of \textsc{D-Attn}~\citep{kuo2025dattn}, we aim to isolate the effect of the attention mechanism rather than pursue state-of-the-art performance through larger-scale data or model scaling. We compare \rave{} with four attention baselines under matched or maximally comparable settings: \textit{Standard Attention}, \textsc{VAR}, \textsc{D-Attn}, and \textsc{Gated-Attn}. The trainable variants use the same language backbone, vision encoder, projector, training data, and optimization schedule, allowing performance differences to be more directly attributed to the attention mechanism. \textit{Standard Attention} retains the original causal self-attention; \textsc{D-Attn} decomposes self-attention into visual-to-visual, textual-to-visual, and textual-to-textual paths; and \textsc{Gated-Attn} applies learned output-side \(O\)-gating. As a training-free method, \textsc{VAR} is applied post hoc to the same trained \textit{Standard Attention} checkpoint without additional training. Detailed construction, implementation, and comparability settings are provided in Appendix~\ref{app:exp-details}.

\paragraph{Benchmarks.}
We evaluate on a suite of multimodal benchmarks covering both perception-intensive and reasoning-oriented capabilities using the VLMEvalKit evaluation framework \citep{duan2024vlmevalkit}. The perception-intensive group includes multilingual OCR, document understanding, chart understanding, and text-rich visual question answering, with MTVQA-TEST~\citep{tang2025mtvqa}, 
ChartQA-TEST~\citep{masry2022chartqa}, 
DocVQA-VAL~\citep{mathew2021docvqa}, 
OCRBench~\citep{liu2023ocrbench}, 
InfoVQA-VAL~\citep{mathew2022infographicvqa}, 
and TextVQA-VAL~\citep{singh2019textvqa}. 
The reasoning-oriented group includes MathVista-MINI~\citep{lu2024mathvista}, 
MMMU-DEV-VAL~\citep{yue2024mmmumassivemultidisciplinemultimodal}, 
and MMMU-Pro-10c~\citep{yue2025mmmupro}, which require visual reasoning, mathematical reasoning, and domain knowledge. This benchmark selection allows us to examine whether visual attention recalibration improves fine-grained visual evidence utilization while preserving general multimodal reasoning ability. For each benchmark, we report the evaluation score produced by VLMEvalKit, together with Perception Avg., Reasoning Avg., and Overall Avg. over the corresponding groups.

\begin{table*}[t]
\centering
\setlength{\tabcolsep}{2.6pt}
\renewcommand{\arraystretch}{1.08}
\resizebox{\textwidth}{!}{
\begin{tabular}{lcccc@{\hspace{8pt}}ccccccccc@{\hspace{6pt}}c}
\toprule
\multirow{2}{*}{Variant}
& \multicolumn{4}{c}{Configuration}
& \multicolumn{10}{c}{Benchmarks} \\
\cmidrule(lr){2-5} \cmidrule(lr){6-15}
& Location
& Form
& $p$
& Application
& MTVQA
& ChartQA
& DocVQA
& OCRBench
& InfoVQA
& TextVQA
& MathVista
& MMMU
& MMMU-Pro
& Avg. \\
\midrule

\multicolumn{15}{l}{\textit{Main configuration}} \\
\midrule
(1)
& Pre.
& Add.
& 0.25
& P+D
& 9.50 & 26.28 & 43.55 & 44.20 & 28.21 & 65.90
& 25.40 & 32.22 & 20.16
& \textbf{32.81} \\

\addlinespace[2pt]
\midrule
\multicolumn{15}{l}{\textit{Location variants }\; (Pre./Post.: pre/post softmax)}  \\
\midrule
(2)
& \textbf{Post.}
& Add.
& 0.25
& P+D
& 8.91 & 24.96 & 44.68 & 43.20 & 28.73 & 65.61 & 24.40 & 32.44 & 19.54 & 32.50 \\

\addlinespace[2pt]
\midrule
\multicolumn{15}{l}{\textit{Form variants}\; (Add./Mul.: additive/multiplicative)} \\
\midrule
(3)
& Pre.
& \textbf{Mul.}
& 0.25
& P+D
& 10.05 & 23.96 & 44.01 & 42.40 & 27.78 & 65.74 & 26.00 & 31.67 & 18.32 & 32.21 \\

\addlinespace[2pt]
\midrule
\multicolumn{15}{l}{\textit{Head ratio variants} \; ($p$: partial-head ratio)} \\
\midrule
(4)
& Pre.
& Add.
& \textbf{0.50}
& P+D
& 10.25 & 23.08 & 42.99 & 41.30 & 28.79 & 65.25 & 26.20 & 32.33 & 19.08 & 32.14 \\
(5)
& Pre.
& Add.
& \textbf{0.75}
& P+D
& 10.37 & 25.00 & 45.21 & 43.60 & 28.78 & 67.13 & 24.40 & 29.00 & 18.61 & 32.46 \\
(6)
& Pre.
& Add.
& \textbf{1.00}
& P+D
& 10.27 & 23.52 & 44.42 & 44.40 & 28.06 & 65.97 & 23.60 & 29.78 & 19.19 & 32.13 \\

\addlinespace[2pt]
\midrule
\multicolumn{15}{l}{\textit{Application variants}\; (P+D/Dec.: prefill+decoding/decoding-only)} \\
\midrule
(7)
& Pre.
& Add.
& 0.25
& \textbf{Dec.}
& 10.15 & 25.28 & 43.36 & 42.70 & 27.98 & 65.28 & 23.10 & 30.78 & 18.79 & 31.94 \\

\bottomrule
\end{tabular}}
\vspace{-5pt}
\caption{
Ablation study of \rave{} design choices across all benchmarks.
Variant (1) is the default \rave{} setting.
Each remaining variant modifies one component of the main configuration while keeping the others unchanged.
The modified component is highlighted in bold.
}
\label{tab:dual:ablation}
\vspace{-10pt}
\end{table*}
\subsection{Main results}

Table~\ref{tab:dual:main-results} summarizes the main results across the evaluated benchmarks. Under the matched Mistral-v0.3-7B setting, \rave{} achieves the best overall average among the compared attention mechanisms. Since the comparisons follow a strictly matched training and evaluation protocol, the observed gains can be more directly attributed to the proposed visual attention recalibration mechanism.

The improvement of \rave{} is most pronounced in perception-intensive scenarios, such as OCR, document understanding, chart understanding, and text-rich VQA, where accurate generation relies heavily on fine-grained visual evidence. This trend aligns with the design of \rave{}: the query-dependent pre-softmax correction to visual-key logits helps the model dynamically modulate both the amount of visual evidence incorporated into generation and the distribution of attention among visual tokens.

On reasoning- and knowledge-heavy benchmarks, such as MathVista, MMMU, and MMMU-Pro, the trend is more nuanced. \rave{} improves over Standard Attention on average, but \textsc{D-Attn} remains stronger on this group. This suggests that \rave{} mainly improves visual evidence allocation without substantially disrupting the reasoning and language-side capabilities inherited from the backbone. Overall, \rave{} provides the clearest benefits in perception-intensive scenarios, where generation relies on fine-grained visual evidence, while preserving competitive performance on broader multimodal reasoning tasks.

The comparisons with \textsc{Gated-Attn} and \textsc{VAR} further clarify the role of the proposed intervention. Under the same multimodal adaptation protocol, \textsc{Gated-Attn} underperforms \rave{} and even falls below the \textit{Standard Attention} baseline, suggesting that adding a learned gate alone is insufficient to explain the gains. This highlights the importance of where and how the gate is applied within the attention computation. In contrast to \textsc{VAR}'s hand-crafted inference-time redistribution, \rave{} employs a learned, content-dependent recalibration mechanism that offers greater flexibility in regulating both the overall attention allocated to visual evidence and its distribution across visual tokens.


\subsection{Ablation studies}

\label{sec:dual:ablations}

Table~\ref{tab:dual:ablation} summarizes the ablation studies of \rave{} on Mistral-v0.3-7B. We examine four aspects of the design: the location and form of attention recalibration, the GQA-aware head ratio, and the application stage. 
Each variant changes one component of the default configuration while keeping the remaining settings unchanged.

\paragraph{Recalibration location and form.}
Pre-softmax recalibration provides a more effective design than post-softmax modulation. By adjusting visual-key logits before normalization, it allows the correction to affect both intra-visual token allocation and cross-modal competition between visual and textual evidence. Additive recalibration is also more stable than its multiplicative counterpart. Although both forms derive recalibration scores from query-key gating, multiplicative modulation introduces a sign-dependent interaction with the original logits, making negative-logit modulation less predictable.


\paragraph{GQA-aware head ratio.}
The head-ratio variants reveal a trade-off between corrective capacity and preservation of the pretrained attention prior. 
The default setting with $p=0.25$ achieves the best overall performance, while larger gated-head ratios do not yield further gains. 
Although gating more heads increases the capacity to modify visual attention, it may also disturb the attention patterns inherited from the pretrained language backbone. 
By leaving part of the query heads ungated within each GQA group, \rave{} preserves standard attention pathways for language-side reasoning and autoregressive generation, while using the gated heads to correct visual evidence allocation. 

\paragraph{Application stage.}
Applying \rave{} during both prefill and decoding outperforms decoding-only application. 
This result suggests that visual attention recalibration is useful not only during answer generation but also during the initial construction of image-conditioned representations. 
By recalibrating question-to-visual attention, the model can form a better visual context before autoregressive decoding begins, which leads to more effective use of visual evidence in subsequent generation.

\subsection{Cross-modal attention allocation}
\label{sec:attention-mass}
To examine whether \rave{} changes the cross-modal allocation pattern identified in Section~\ref{sec:dual:motivation}, we compare the segment-wise image attention mass of the Mistral-based \textit{Standard Attention} and \rave{} models on the same perception and reasoning probes. As shown in Table~\ref{tab:attention-mass}, \rave{} consistently assigns more attention mass to the image segment at both the first and final decoding positions across the two task groups. In particular, it maintains substantially greater visual attention at the final decoding position, indicating that \rave{} mitigates cross-modal visual under-allocation late in generation. Full sampling details and statistics are provided in Appendix~\ref{app:exp-details}.

\vspace{-1pt}
\subsection{Backbone generalization}
\vspace{-1pt}
To examine whether \rave{} transfers beyond the controlled LLaVA-style setting, we additionally evaluate it on Qwen2.5-VL-7B-Instruct~\citep{bai2025qwen25vltechnicalreport}, a multimodal model that has already undergone multimodal pretraining and instruction tuning. Starting from the same released checkpoint, the \textit{Standard Attention} and \rave{} variants undergo matched continued supervised
fine-tuning. Full settings are provided in Appendix~\ref{app:exp-details}. As shown in Table~\ref{tab:dual:qwen-results}, \rave{} achieves a higher average score than \textit{Standard Attention} across the nine benchmarks, supporting the transferability of the proposed recalibration to an already instruction-tuned LMM. The improvement is more modest than in the Mistral-based setting, possibly because Qwen2.5-VL already has well-established multimodal alignment and attention patterns before \rave{} is introduced. Continued fine-tuning may therefore provide less opportunity for the newly inserted gate to co-adapt with and substantially reshape the decoder attention, while the stronger starting model also leaves less headroom for improvement.

\vspace{-1pt}
\section{Conclusion}
\vspace{-1pt}
\label{conclusion}
We presented \rave{}, a lightweight attention recalibration mechanism for decoder-only large multimodal models. Motivated by the observation that standard attention can misallocate visual evidence across both modality-level and intra-visual dimensions, \rave{} introduces a query-dependent pre-softmax correction to attention logits over visual keys. This enables context-aware visual attention reallocation without modifying the backbone architecture or changing the standard training pipeline. Under matched model and training configurations, \rave{} achieves clear improvements on multimodal benchmarks, with the strongest gains on perception-intensive tasks that require fine-grained visual grounding, while maintaining competitive performance on reasoning- and knowledge-heavy benchmarks. Ablation studies show that pre-softmax additive recalibration, GQA-aware partial-head intervention, and prefill+decoding application are all important to the final design. These results suggest that visual attention recalibration is an effective mechanism-level correction for improving visual evidence utilization in decoder-only LMMs. 

\vspace{-1pt}
\section*{Limitations}
\vspace{-1pt}

Our experiments focus on instruction-tuned LMMs built from pretrained language backbones. Due to the substantial computational cost of training LMMs from scratch or pretraining native multimodal backbones, we do not evaluate \rave{} under such settings. Following the controlled protocol of \textsc{D-Attn}, we restrict our main controlled comparison to a matched decoder-only LMM setup, which helps isolate the effect of the attention mechanism. Although we additionally validate \rave{} on Qwen2.5-VL-7B-Instruct, evaluation across broader model scales and multimodal architectures remains future work. In addition, our current implementation is based on SDPA and has not yet been integrated with FlashAttention-style kernels, which may affect training and inference efficiency.

\section*{Ethical Considerations}

This work studies attention recalibration for large multimodal models and does not collect, create, or release new datasets containing private, sensitive, or personally identifiable information. It also does not involve human participants or annotators. We use publicly available models, datasets, and benchmarks, with citations to their original creators. The potential risks are similar to those of general LMM systems, including misuse, over-reliance, and insufficiently supervised deployment. Although \rave{} improves visual evidence utilization, it does not eliminate hallucination or guarantee reliability in high-stakes settings. We recommend careful evaluation, appropriate safeguards, and human oversight for real-world use.

\section*{Acknowledgments}

This work was supported by Qwen Business Unit through Alibaba Research Intern Program.

\bibliography{ref}


\appendix

\section{Experimental Details}

\label{app:exp-details}
\paragraph{Controlled evaluation protocol.}
Our experiments comprise two complementary settings: a primary controlled comparison built on Mistral-v0.3-7B \citep{jiang2023mistral7b}, and a separate backbone-generalization study on Qwen2.5-VL-7B-Instruct \citep{bai2025qwen25vltechnicalreport}.
The primary comparison follows the controlled evaluation protocol of \textsc{D-Attn}~\citep{kuo2025dattn}. The goal is not to maximize absolute performance through larger-scale data, stronger backbones, or additional alignment stages, but to isolate the effect of the attention mechanism. Accordingly, the trainable Mistral-based variants---\textit{Standard Attention}, \textsc{D-Attn}, \textsc{Gated-Attn}, and \rave{}---use the same pretrained language backbone, visual encoder, projector, training
data, and optimization schedule. The only intended difference among these variants is the attention intervention used inside the decoder. Because \textsc{VAR} is a training-free inference-time method, we apply it
post hoc to the same trained \textit{Standard Attention} checkpoint rather than training a separate model.

\paragraph{Scope of comparison.} Table~\ref{tab:dual:positioning} organizes prior approaches to visual-attention allocation into three broad families: training-free inference-time intervention, architectural redesign, and training-based visual revisiting. Our numerical comparison includes representative methods from the first two families that can be evaluated under matched or maximally comparable conditions. For training-free intervention, we select \textsc{VAR} because it directly targets visual attention sinks, a central form of intra-visual misallocation considered in this work. For architectural redesign, we select \textsc{D-Attn} because it provides the closest architectural counterpart: it modifies visual--text interaction by decomposing attention into modality-specific paths, whereas \rave{} retains native joint attention and recalibrates the pre-softmax logits over visual keys. We additionally include \textsc{Gated-Attn} as a mechanism-level comparator because, like \rave{}, it introduces learned gating within the attention operator, while applying the gate to the attention output rather than to the attention allocation. We do not include training-based revisiting methods such as \textsc{TVC} and \textsc{Look-Back} as numerical baselines because their original pipelines rely on curated long-chain reasoning data and specialized visual-revisiting generation procedures. \textsc{Look-Back} further uses supervised fine-tuning, reinforcement learning, and a dedicated visual-refocusing token. Including these methods would therefore alter both the training data and the decoding protocol, preventing a controlled comparison focused on the attention mechanism.

\paragraph{Mistral-based model construction and baselines.}
Following \textsc{D-Attn}, we adopt a LLaVA-style decoder-only LMM architecture consisting of a pretrained SigLIP~\citep{zhai2023sigmoid} visual encoder, a two-layer MLP projector with RMSNorm for visual--language alignment, and a pretrained Mistral-v0.3-7B language backbone. Under this architecture, \textit{Standard Attention} retains the original causal self-attention mechanism. \textsc{D-Attn} decomposes causal attention into visual-to-visual, textual-to-visual, and textual-to-textual paths. For \textsc{Gated-Attn}~\citep{qiu2025gatedattention}, we follow the original paper's best-performing \(O\)-gating configuration, which applies a head-specific sigmoid gate to the attention output. We use its reported hyperparameters where applicable, without
method-specific tuning on our evaluation benchmarks. \rave{} retains the native joint-attention formulation while adding the proposed query--key pair gate to the pre-softmax logits associated with visual keys. For the training-free comparison, we apply \textsc{VAR}~\citep{kang2025visualsink} to the trained \textit{Standard Attention} model. Specifically, we re-identify the sink dimensions on our Mistral-based checkpoint following the procedure in the original work, and then apply its attention-redistribution rule during inference using the default hyperparameters.


\paragraph{Training recipe.} The trainable Mistral-based variants follow the three-stage training recipe of \textsc{D-Attn} where applicable. In the first stage, the visual--language projector is trained for modality alignment using the LLaVA LAION/CC/SBU \citep{liu2023visual,ordonez2011im2text,schuhmann2022laion5bopenlargescaledataset,sharma2018conceptual} 558K subset. In the second stage, the model is fine-tuned on 1.25M dense captions from ShareGPT4V-PT~\citep{chen2024sharegpt4v}. In the third stage, instruction tuning is performed using a combination of 665K LLaVA-1.5 instruction-tuning samples~\citep{liu2023visual} and 102K dense-caption samples from ShareGPT4V~\citep{chen2024sharegpt4v}. All trainable variants use the standard next-token cross-entropy objective. For \textsc{Gated-Attn} and \rave{}, the added gate parameters are optimized jointly with the LMM during the stages in which the decoder is updated. \rave{} requires neither auxiliary supervision nor an additional gating-specific loss. \textsc{VAR} requires no additional training. For the Qwen2.5-VL experiment, both matched variants undergo continued supervised fine-tuning using the datasets from all three stages described above, starting from the same publicly released instruction-tuned checkpoint.

\paragraph{Evaluation protocol.} The original \textsc{D-Attn} study evaluates ten image benchmarks under the LLaVA evaluation protocol. In this work, we instead use VLMEvalKit~\citep{duan2024vlmevalkit} to evaluate a benchmark suite covering both perception-intensive and reasoning-oriented capabilities. The perception-intensive group includes MTVQA-TEST, ChartQA-TEST, DocVQA-VAL, OCRBench, InfoVQA-VAL, and TextVQA-VAL. The reasoning-oriented group includes MathVista-MINI, MMMU-DEV-VAL, and MMMU-Pro-10c. For each benchmark, we report the score returned by VLMEvalKit, together with \textit{Perception Avg.}, \textit{Reasoning Avg.}, and \textit{Overall Avg.} over the corresponding benchmark groups. The same benchmark definitions and evaluation settings are used for both the Mistral-based comparison and the Qwen2.5-VL backbone study.

\paragraph{Cross-modal attention-mass analysis.} To examine whether \rave{} changes the cross-modal allocation pattern identified in Section~\ref{sec:dual:motivation}, we compare the Mistral-based \textit{Standard Attention} and \rave{} models on the same perception and reasoning probes. For the perception group, we randomly sample 10 examples each from OCRBench and DocVQA; for the reasoning group, we randomly sample 10 examples each from MathVista and MMMU, yielding 20 probes per group. Both models use greedy decoding with identical images, image preprocessing, and prompt templates. Following Eq.~\ref{eq:dual:vis-mass}, we compute the image- and answer-segment attention masses at each generated position and average them over decoder layers, query heads, and probe examples. Table~\ref{tab:attention-mass} reports the first and final decoding positions, for which all 20 probes in each group contribute. Across both task groups, \rave{} assigns substantially more attention to the image segment, including at the final decoding position. These results indicate that \rave{} mitigates cross-modal visual under-allocation late in generation.
\begin{table}[t]
\vspace{5pt}
\centering
\setlength{\tabcolsep}{2.8pt}
\renewcommand{\arraystretch}{1.08}
\resizebox{\columnwidth}{!}{
\begin{tabular}{lcccc|cc}
\toprule
\multirow{2}{*}{Group}
& \multirow{2}{*}{Position}
& \multicolumn{3}{c|}{Image Mass (\%)}
& \multicolumn{2}{c}{Answer Mass (\%)} \\
\cmidrule(lr){3-5}
\cmidrule(lr){6-7}
& & Std. & \rave{} & $\Delta$ & Std. & \rave{} \\
\midrule
\multirow{2}{*}{Perception}
& First & 24.8 & 39.6 & +14.8 & 4.8  & 4.1  \\
& Final  & 20.9 & 34.7 & +13.8 & 16.0 & 15.5 \\
\midrule
\multirow{2}{*}{Reasoning}
& First & 12.8 & 26.1 & +13.3 & 5.0 & 4.3 \\
& Final  & 12.3 & 25.3 & +13.0 & 7.2 & 6.1 \\
\bottomrule
\end{tabular}
}
\vspace{-5pt}
\caption{
Segment-wise attention mass (\%) for \textit{Standard Attention} and
\rave{} at the first and final decoding positions.
Values are averaged over all decoder layers, query heads, and 20 probes
per task group. $\Delta$ denotes the image-segment attention mass of
\rave{} minus that of \textit{Standard Attention}.
}
\vspace{-10pt}
\label{tab:attention-mass}
\end{table}

\begin{table*}[t]
\centering
\setlength{\tabcolsep}{3.2pt}
\renewcommand{\arraystretch}{1.08}
\resizebox{\textwidth}{!}{
\begin{tabular}{lccccccccc|ccc}
\toprule
 \multirow{2}{*}{Attention Method}
& \multicolumn{6}{c}{Perception-intensive}
& \multicolumn{3}{c}{Reasoning / Knowledge-heavy}
& \multirow{2}{*}{Perc. Avg.}
& \multirow{2}{*}{Reason. Avg.}
& \multirow{2}{*}{Overall Avg.} \\
\cmidrule(lr){2-7}
\cmidrule(lr){8-10}

& MTVQA
& ChartQA
& DocVQA
& OCRBench
& InfoVQA
& TextVQA
& MathVista
& MMMU
& MMMU-Pro
& & & \\
\midrule
 \textit{Standard Attention}
& 22.03 & 80.04 & 92.33 & 82.50 & 74.76 & 82.38
& 58.20 & 48.11 & 35.18 & 72.34 & 47.16 & 63.95 \\
\rave{}
& \textbf{22.19} & \textbf{82.44} & \textbf{92.85} & \textbf{84.40} & \textbf{76.10} & \textbf{82.96}
& \textbf{58.50} & \textbf{49.22} & \textbf{36.06}
& \textbf{73.49} & \textbf{47.93} & \textbf{64.97} \\

\bottomrule
\end{tabular}}
\vspace{-5pt}
\caption{
Backbone-generalization results on Qwen2.5-VL-7B-Instruct. Both variants start from the same publicly released instruction-tuned checkpoint and undergo matched continued supervised fine-tuning using the datasets from all three stages of our controlled recipe; they differ only in whether \rave{} is inserted into the language decoder.
}
\label{tab:dual:qwen-results}
\vspace{-10pt}
\end{table*}

\balance
\paragraph{Backbone-generalization setting.} To assess whether \rave{} transfers beyond the controlled LLaVA-style architecture, we additionally use the publicly released Qwen2.5-VL-7B-Instruct checkpoint~\citep{bai2025qwen25vltechnicalreport}. Unlike the Mistral-based models constructed from a pretrained language backbone through our multimodal training pipeline, Qwen2.5-VL-7B-Instruct is already an instruction-tuned multimodal model with its own vision tower and M-RoPE-based positional treatment. We therefore treat this experiment as continued supervised fine-tuning from an existing instruction-tuned checkpoint, rather than as a reproduction of Qwen2.5-VL's original multimodal pretraining and instruction-tuning pipeline. Starting from the same released checkpoint, the \textit{Standard Attention} and \rave{} variants undergo continued supervised fine-tuning using the data from all three stages of our controlled recipe: the 558K LLaVA LAION/CC/SBU subset used in Stage 1, the 1.25M ShareGPT4V-PT dense-caption set used in Stage 2, and the instruction-tuning mixture used in Stage 3. The two variants use the same data sequence, training schedule, and optimization settings; the only difference is whether \rave{} is inserted into the language decoder.


\paragraph{Computational overhead.} In the matched Qwen2.5-VL continued-SFT experiment under the same compute configuration, \textit{Standard Attention} and \rave{} require approximately 20 and 22 hours, respectively, for the same training schedule, corresponding to approximately \(10\%\) additional training time. Under the same compute configuration, batch settings, and maximum generation lengths in VLMEvalKit, evaluation on the six perception-intensive benchmarks takes approximately 12 hours for \textit{Standard Attention} and 14 hours for \rave{}, corresponding to approximately \(17\%\) end-to-end inference overhead. These measurements reflect our current non-fused SDPA implementation, which explicitly materializes the pairwise gate tensor, and therefore provide a conservative estimate of the deployment overhead. Because each gate entry is generated from two per-token scalar vectors, a fused implementation could compute the correction on the fly without materializing an additional \(N \times N\) gate tensor.

\paragraph{Baseline implementation and reproducibility.} For \textsc{D-Attn}, we use the official implementation and reported hyperparameter settings without additional tuning. For \textsc{Gated-Attn}, we reproduce the original paper's best-performing output-side \(O\)-gating configuration. We note that the original \textsc{Gated-Attn} study introduces its gate during language-model pretraining, whereas our controlled setup introduces it during multimodal adaptation, consistent with the other trainable Mistral-based variants. Thus, this comparison evaluates \textsc{Gated-Attn} as a drop-in intervention under a matched multimodal training budget rather than reproducing its full language-model pretraining setting. For \textsc{VAR}, we follow the original inference-time procedure and default hyperparameters, while identifying the sink dimensions on our own trained backbone. For the Qwen2.5-VL study, the two variants share the same initial checkpoint, training data, and optimization schedule, and differ only in whether \rave{} is inserted into the language decoder. We use publicly released models, datasets, and benchmarks for research purposes and follow their released licenses or terms of use. Our implementation is based on PyTorch SDPA, DeepSpeed, and VLMEvalKit. Package versions and scripts will be released with the code.

\end{document}